\documentclass{article} 
\usepackage{iclr2025_ai4na_tiny_paper, times}
\usepackage{amsmath} 

\usepackage{amsmath,amsfonts,bm}

\def\eqref#1{equation~\ref{#1}}

\def\1{\bm{1}}

\DeclareMathAlphabet{\mathsfit}{\encodingdefault}{\sfdefault}{m}{sl}
\SetMathAlphabet{\mathsfit}{bold}{\encodingdefault}{\sfdefault}{bx}{n}

\usepackage{hyperref}
\usepackage{url}
\usepackage{graphicx}

\title{Transfer Orthology Networks}

\author{Vikash Singh\\
Stillmark\\
\texttt{vikash@stillmark.com}
}

\iclrfinalcopy 
\begin{document}

\maketitle

\begin{abstract}
 We present Transfer Orthology Networks (TRON), a novel neural network architecture designed for cross-species transfer learning. TRON leverages orthologous relationships, represented as a bipartite graph between species, to guide knowledge transfer.  Specifically, we prepend a learned species conversion layer, whose weights are masked by the biadjacency matrix of this bipartite graph, to a pre-trained feedforward neural network that predicts a phenotype from gene expression data in a source species.  This allows for efficient transfer of knowledge to a target species by learning a linear transformation that maps gene expression from the source to the target species' gene space.  The learned weights of this conversion layer offer a potential avenue for interpreting functional orthology, providing insights into how genes across species contribute to the phenotype of interest.  TRON offers a biologically grounded and interpretable approach to cross-species transfer learning, paving the way for more effective utilization of available transcriptomic data. We are in the process of collecting cross-species transcriptomic/phenotypic data to gain experimental validation of the TRON architecture. 
\end{abstract}

\section{Related Work}

Recent advancements in cross-species transfer learning have significantly improved our ability to translate biological findings across different organisms. Species-Agnostic Transfer Learning (SATL) has emerged as a novel approach that allows knowledge transfer beyond species barriers without relying on known gene orthology \cite{sun2024species}. This method utilizes heterogeneous domain adaptation and extends the cross-domain structure-preserving projection for out-of-sample prediction. Other notable contributions include GDEC, which effectively clusters cross-species and cross-batch scRNA-seq data \cite{sun2023transfer}, and FloraBERT, a deep learning model that improves gene expression predictions by exploiting cross-species genomic information in plants \cite{wang2024florabert}. Computational strategies for cross-species knowledge transfer have also been developed, focusing on transcriptome data and molecular networks \cite{li2024computational}. These approaches address challenges such as transferring disease and gene annotation knowledge, identifying analogous molecular components, and mapping cell types across species. Additionally, unsupervised domain adaptation methods have been applied to transfer learning tasks involving genomic regulatory code signals between different species \cite{wang2024unsupervised}.

\section{Methodology}

\subsection{Constructing the Bipartite Graph of Orthologous Relationships}

The core of TRON's species conversion layer lies in the bipartite graph $B$, which encodes the orthologous relationships between genes in the source and target species.  This graph serves as a mask, guiding the knowledge transfer process by restricting the connections to only those between orthologous gene pairs.  

In this work, we focus on constructing $B$ using reciprocal best hits (RBHs). RBHs are identified by performing all-against-all sequence comparisons (e.g., using BLAST or DIAMOND) between the gene sets of the two species. A gene $j$ in the source species is considered a "best hit" for gene $i$ in the target species if it has a sequence similarity score above a set threshold and is among the highest scoring genes when compared to gene $i$. Depending on the chosen threshold, it is possible for multiple genes $j_1, j_2, ...$ in the source species to be considered best hits for gene $i$. Reciprocally, for each of these genes $j_1, j_2, ...$, gene $i$ in the target species must also be their respective best hit, above the same threshold, in the target species. If these reciprocal best hit conditions are met, the gene pairs $(i, j_1), (i, j_2), ...$ are each considered reciprocal best hits, indicating a strong likelihood of orthology. This allows for a single target gene to have multiple orthologous counterparts in the source species, or vice versa, resulting in a many-to-many orthologous relationship.

The biadjacency matrix $B$ is then constructed based on these RBHs.  Specifically:

$$
B_{ij} =
\begin{cases}
1, & \text{if genes } i \text{ and } j \text{ are reciprocal best hits} \\
0, & \text{otherwise}
\end{cases}
$$

 Alternative orthology detection methods, such as phylogenetic tree-based approaches, could be explored in future work to potentially improve the accuracy and completeness of the bipartite graph and, consequently, the performance of TRON.  Investigating the impact of different orthology calling methods on TRON's performance is an interesting avenue for future research.
\begin{figure}[!h]
\begin{center}
\includegraphics[width=0.8\textwidth]{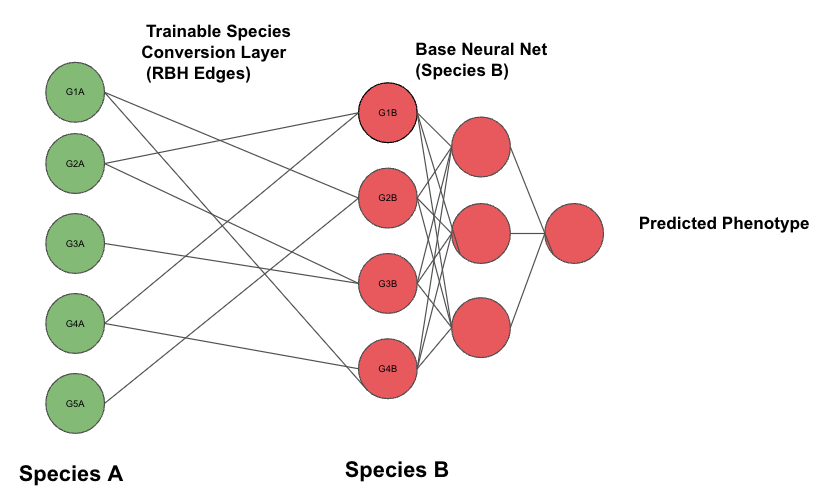}
\end{center}
\caption{TRON Architecture}
\end{figure}

\subsection{Species Conversion Layer}

Let:

 $x_s \in \mathbb{R}^{n_s}$ be the gene expression vector of the source species, where $n_s$ is the number of genes in the source species.
 $W_c \in \mathbb{R}^{n_t \times n_s}$ be the weight matrix of the conversion layer, where $n_t$ is the number of genes in the target species.
 $B \in \{0, 1\}^{n_t \times n_s}$ be the biadjacency matrix of the bipartite graph representing orthologous relationships between the source and target species. $B_{ij} = 1$ if gene $j$ in the source species is orthologous to gene $i$ in the target species, and $0$ otherwise.
 $\odot$ denote the Hadamard product (element-wise multiplication).

The output of the species conversion layer, $x_t \in \mathbb{R}^{n_t}$, is computed as:

$$x_t = (W_c \odot B) x_s$$

This can be rewritten element-wise as:

$$x_{t_i} = \sum_{j=1}^{n_s} (W_{c_{ij}} \cdot B_{ij}) x_{s_j}, \quad i = 1, 2, ..., n_t$$

Where:

 $x_{t_i}$ is the expression level of gene $i$ in the target species.
 $W_{c_{ij}}$ is the weight associated with the connection between gene $j$ in the source species and gene $i$ in the target species.
 $B_{ij}$ is the binary indicator of whether gene $j$ in the source species is orthologous to gene $i$ in the target species.
 $x_{s_j}$ is the expression level of gene $j$ in the source species.

The masked weight matrix, $W_c \odot B$, ensures that only connections corresponding to orthologous gene pairs contribute to the transformed gene expression vector $x_t$. Weights corresponding to non-orthologous pairs are effectively zeroed out.

Prepend to the original neural network:

Let $f(\cdot)$ be the original neural network trained on the source species data. The input to $f(\cdot)$ is now $x_t$:

$$y = f(x_t) = f((W_c \odot B) x_s)$$

where $y$ is the predicted phenotype.

During training, the weights of the original neural network $f(\cdot)$ are frozen, and only the weights of the conversion layer, $W_c$, are updated using gradient descent.

\subsection{Functional Orthology and Interpretability}

The training process of the species conversion layer offers a unique perspective on functional orthology.  As only the weights $W_c$ of the conversion layer are learned while the original network $f(\cdot)$ is frozen, the conversion layer effectively learns a linear combination of the source species gene expression to approximate the target species gene expression.  Crucially, due to the masking by the biadjacency matrix $B$, this linear combination only considers the contributions of orthologous genes.

Specifically, for each gene $i$ in the target species, its expression level $x_{t_i}$ is computed as a weighted sum of the expression levels of its orthologs in the source species:

$$x_{t_i} = \sum_{j=1}^{n} (W_{c_{ij}} \cdot B_{ij}) x_{s_j}, \quad i = 1, 2, ..., n_t$$

Since $B_{ij}$ is binary, $W_{c_{ij}} \cdot B_{ij}$ is zero if genes $i$ and $j$ are not orthologous. Thus, we can rewrite this as:

$$x_{t_i} = \sum_{j \in \text{orthologs of } i} W_{c_{ij}} x_{s_j}, \quad i = 1, 2, ..., n_t$$

The weights $W_{c_{ij}}$ thus represent the relative contribution of each orthologous gene $j$ in the source species to the expression level of gene $i$ in the target species. These learned weights can be interpreted as reflecting the functional similarity or relationship between the orthologous gene pair.  A larger magnitude of $W_{c_{ij}}$ suggests a stronger influence of gene $j$ on the expression of gene $i$, potentially indicating a closer functional relationship.  Analyzing these weights can provide insights into the functional divergence or conservation of orthologous genes across species, offering a valuable lens into functional orthology. This interpretability is a significant advantage of the TRON architecture and opens avenues for exploring the functional implications of cross-species gene expression relationships.

\appendix

\section{Appendix}

\subsection{Soft Orthology Constraint via Regularization}

To introduce a soft constraint on the orthologous relationships within the species conversion layer, we modify the loss function by adding regularization terms based on the biadjacency matrix $B$.  Let $L_0$ be the original loss function used to train the species conversion layer.  We introduce two additional terms: one for the weights corresponding to orthologous genes and another for the weights corresponding to non-orthologous genes.

Let:

 $W_c \in \mathbb{R}^{m \times n}$ be the weight matrix of the species conversion layer.
 $B \in \{0, 1\}^{m \times n}$ be the biadjacency matrix of the bipartite graph representing orthologous relationships.
 $\alpha$ and $\beta$ be non-negative hyperparameters controlling the strength of the regularization.

We define the modified loss function $L$ as:

$$L = L_0 + \alpha \sum_{i=1}^{m} \sum_{j=1}^{n} (1 - B_{ij}) W_{c_{ij}}^2 + \beta \sum_{i=1}^{m} \sum_{j=1}^{n} B_{ij} W_{c_{ij}}^2$$

The first regularization term, $\alpha \sum_{i=1}^{m} \sum_{j=1}^{n} (1 - B_{ij}) W_{c_{ij}}^2$, penalizes the weights $W_{c_{ij}}$ when genes $i$ and $j$ are not orthologous (i.e., $B_{ij} = 0$). The term $(1-B_{ij})$ acts as a mask that selects the non-orthologous pairs. The hyperparameter $\alpha$ controls the strength of this penalty. A larger $\alpha$ enforces a stronger preference for zero weights for non-orthologous connections.

The second regularization term, $\beta \sum_{i=1}^{m} \sum_{j=1}^{n} B_{ij} W_{c_{ij}}^2$, penalizes the weights $W_{c_{ij}}$ when genes $i$ and $j$ are orthologous (i.e., $B_{ij} = 1$). The term $B_{ij}$ acts as a mask that selects the orthologous pairs. The hyperparameter $\beta$ controls the strength of this penalty. A larger $\beta$ encourages smaller weights for orthologous connections.

By adjusting $\alpha$ and $\beta$, we can control the degree to which the orthologous relationships are enforced.  Setting $\alpha \gg \beta$ would strongly penalize non-orthologous connections, effectively approximating a hard constraint.  Conversely, setting $\alpha \approx \beta$ would allow for more flexibility, permitting non-zero weights for non-orthologous connections if they contribute to improved prediction performance.  Setting $\beta$ to a very small value or zero allows the model to learn the strength of the orthologous connections.

This soft constraint allows the model to learn potentially valuable connections beyond strict orthology while still leveraging the prior knowledge encoded in the bipartite graph.

\bibliography{iclr2025_ai4na_tiny_paper}
\bibliographystyle{iclr2025_conference}


\end{document}